\documentclass[10pt,twocolumn,letterpaper]{article}

\usepackage{iccv}
\usepackage{times}
\usepackage{epsfig}
\usepackage{graphicx}
\usepackage{amsmath}
\usepackage{multirow}
\usepackage{amssymb}
\usepackage{authblk}

\usepackage[breaklinks=true,bookmarks=false]{hyperref}

\iccvfinalcopy 

\def\iccvPaperID{****} 

\ificcvfinal\fi

\begin{document}

\title{Image Segmentation using U-Net Architecture for Powder X-ray Diffraction Images}


\author[1]{Howard Yanxon}
\author[2]{Eric Roberts}
\author[1]{Hannah Parraga}
\author[1]{James Weng}
\author[1]{Wenqian Xu}
\author[1]{Uta Ruett}
\author[2]{Alexander Hexemer}
\author[2]{Petrus Zwart}
\author[1]{Nicholas Schwarz}

\affil[1]{Advanced Photon Source (APS), Argonne National Laboratory; Lemont, IL 60439}
\affil[2]{Advanced Light Source (ALS), Lawrence Berkeley National Laboratory; Berkeley, CA 94720}

\maketitle
\ificcvfinal\thispagestyle{empty}\fi

\begin{abstract}
    Scientific researchers frequently use the in situ synchrotron high-energy powder X-ray diffraction (XRD) technique to examine the crystallographic structures of materials in functional devices such as rechargeable battery materials.  We propose a method for identifying artifacts in experimental XRD images. The proposed method uses deep learning convolutional neural network architectures, such as tunable U-Nets to identify the artifacts. In particular, the predicted artifacts are evaluated against the corresponding ground truth (manually implemented) using the overall true positive rate or recall. The result demonstrates that the U-Nets can consistently produce great recall performance at 92.4\% on the test dataset, which is not included in the training, with a 34\% reduction in average false positives in comparison to the conventional method. The U-Nets also reduce the time required to identify and separate artifacts by more than 50\%. Furthermore, the exclusion of the artifacts shows major changes in the integrated 1D XRD pattern, enhancing further analysis of the post-processing XRD data.
\end{abstract}

\section{Introduction}

Synchrotron X-ray user facilities produce vast datasets with high brilliance and collimation of synchrotron radiation, offering a wide energy spectrum with tens of thousands of times more photon flux than typical clinical setting \cite{balerna2015synchrotron, winick2012synchrotron}. In these facilities, the experiments encompass a broad range of multidisciplinary sciences and produce data that often comprise a combination of various materials in the X-ray beam path. Identifying relevant data, ensuring adequate data quality, and identifying systematic errors necessitate a high level of expertise. Therefore, coordinating data collection and analysis is crucial for optimizing scientific research efforts. Efforts have focused on optimizing data acquisition processes to enhance scientific reproducibility, transparency, and reliability to overcome limitations in data quality for new discoveries. Successful implementation of these efforts will enable users to efficiently search and comprehend the complex concepts of natural phenomena.

X-ray diffraction (XRD) is a highly utilized technique in synchrotron X-ray facilities. XRD is a non-destructive and highly sensitive method for characterizing crystalline materials in inorganic materials. The method enables researchers to obtain detailed information about the identities and structures of the phases in their samples without altering or damaging them. Moreover, sample particle size, strain, structural defects, and other structural and compositional information can also be extracted from XRD data, making XRD an important tool for both academia and many industries including pharmaceutical, chemical, energy storage, and electronics.

In the pharmaceutical industry, XRD has been used to quantify the presence of specific phases or chemical components in drug samples. XRD analysis allows for the characterization of the crystal structures of drugs and their intermediates, enabling improvement of the physical and chemical properties of drugs \cite{datta2004crystal, halasz2013real}. In the energy storage industry, XRD has been used to study the microstructure of advanced materials, such as lithium-ion batteries. By understanding the crystal structure and the response of the structures of the cathode/anode materials in the charge-discharge process, researchers can improve the performance and stability of energy storage systems \cite{chan2008high, jeong2020microclusters}. In the electronics industry, XRD has been used to study the structure of thin films \cite{hammad2018structural} and other complex materials that are widely used in optical applications such as optical fibers \cite{ballato2008silicon} and solar cells \cite{schelhas2016monitoring}. XRD can also be used to determine the degree of crystallinity in amorphous materials, such as polymers, and their optical properties, such as refractive index and optical bandgap \cite{kumar2013structure}. By using XRD, researchers can gain insights into the properties and behavior of materials, which can lead to the development of new technologies.

With the advent of photon-detecting systems and high photon flux, many synchrotron XRD instruments use large-area 2D detectors to capture diffracted X-rays passing through the sample. Compared to the traditional point detector system, these 2D detectors can achieve sub-second temporal resolution. This higher frequency of data collection is crucial for understanding the transformation of materials under conditions relevant to their applications. In situ or operando experiments, which involve recording a stream of XRD images while the sample undergoes physical or chemical processes, account for more than 90\% of experiments performed at synchrotron XRD beamlines. Researchers from various fields take advantage of advanced source and detector hardware to collect large datasets. However, processing such a high volume of data in a relatively short period of time, performing pre-analysis, preparing the data for more advanced analysis, and presenting it in a way that is easy to understand, are all critical in the big data era.

In this paper, we will demonstrate the capability of convolutional neural network (CNN) methods based on U-Net architectures for recognizing and segmenting artifacts that appear in typical XRD images. XRD images of non-ideal powder samples, which is often the case for many in situ experiments, are usually composed of many complicated characteristics such as diffraction rings, preferred orientation or texture rings, and single-crystal diffraction spots. In many situations, it is desirable to remove single-crystal diffraction spots \cite{yanxon2023artifact} before the integration procedure in order to produce the conventional 1D XRD pattern with accurate peak intensities and profiles. Software such as GSAS--II \cite{toby2013gsas}---state-of-the-art software for crystallographic structural analysis---has a segmentation algorithm that can automatically identify single-crystal diffraction spots based on image pixel intensity. Nevertheless, the algorithm can fail to differentiate the artifacts when other characteristics, such as preferred orientation, are also present in an XRD image.

The performances of the U-Net-based CNN method are compared based on their abilities to identify single-crystal diffraction spots. The demonstration will show suitable approaches to recognize and separate (i.e. mask) in accuracy despite a number of confounding factors that may impact the accuracy of results, such as ring shifting due to crystal structure expansion or contraction. In addition, the speed-up makes the CNN method applicable for on-the-fly masking during an XRD experiment. This implementation not only reduces the huge efforts required to remove artifacts but also enables precise on-the-fly data analysis optimizing data collection for advancing materials discovery.


\section{Related Work}

Convolutional neural networks (CNNs) are deep learning architectures composed of several connected convolutional layers that approximate an unknown function mapping input data to a target domain \cite{drozdzal2016importance, lecun1998gradient}. In this study, the learning is supervised (i.e. the network output is scored against the known ground truth) and the target domain is the binary classification of each pixel to a specific label (e.g. mask or no mask). Each convolutional layer convolves the output of the preceding layer to create an intermediate feature map. The typical CNN learning process relies on an increasing number of two-dimensional convolutional filters, each containing weights that are optimized during training, that represent a learned feature that can be applied to different spatial coordinates (translation equivariance) and more complex image reconstruction tasks in downstream network layers. Between adjacent convolutional layers, nonlinear activation functions and normalization layers are used to expedite the learning process, and max-pooling operations are used to introduce translation invariance and reduce computational costs through spatial coordinate downsampling. This allows for a more global and efficient learning network architectures than traditional, non-convolutional, fully-connected neural networks (FCNN) \cite{goodfellow2016deep, rosenblatt1958perceptron}, where learned features remain localized to the single spatial coordinates.

In this study, we employ U-Nets \cite{ronneberger2015u}, a popular deep CNN architecture that has established itself as a valuable tool in image segmentation due to its versatility and effectiveness. U-Nets are particularly useful when annotated data is scarce, as it has shown to handle complex segmentation tasks in medical imaging with limited training data \cite{bardis2020deep, huang2020machine, nofallah2022segmenting, cciccek20163d}, yielding significant improvements for image segmentation analysis and computer-aided diagnosis across a multitude of domains and applications. Its unique design, consisting of a contracting path for feature capture and a typically symmetric expanding path for image localization awareness, makes it well-suited for use in various imaging applications such as 3D volumetric segmentation \cite{cciccek20163d}, photoacoustic tomography \cite{guan2019fully}, and artifact reduction \cite{hegazy2019u}. The versatility of the U-Net makes it well-suited for various anomaly detection applications, as it can be trained on different datasets with various levels of complexity. For example, in the case of crack detection in concrete surfaces \cite{liu2019computer}, the U-Net can learn to distinguish cracks from other structures in the image, and accurately locate them. Similarly, in road segmentation \cite{abderrahim2020road}, the network can learn to distinguish roads from other structures, such as buildings or vegetation. This makes it a useful resource for researchers and practitioners in a wide range of fields, including artifact identification in XRD images. Subsequently, the U-Net architecture's ability to provide fast and accurate results aligns well with the requirements for artifact removal in XRD images.

\begin{table}[htp]
\begin{center}
\begin{tabular}{|l|c|c|c|}
\hline
 Dataset           & No. of Images & Characteristics            \\
\hline \hline
 Nickel83       & 11            & SCD spots                     \\
 Battery-1      & 11            & SCD spots and POs             \\
 Battery-2      & 12            & SCD spots and textures        \\
 Battery-3      & 12            & SCD spots and textures        \\
 Battery-4      & 14            & SCD spots and textures        \\
 Battery-5      & 12            & SCD spots and textures        \\
 Perfect        & 27            & Only perfect powder rings     \\
\hline
\end{tabular}
\end{center}
\caption{Characteristics of datasets. SCD spots and POs stand for single-crystal diffraction spots and preferred orientation, respectively. Note: all images contain perfect powder rings.}
\label{datasets_table}
\end{table}

\section{Method}
\subsection{Dataset}

\begin{figure}[tp]
\begin{center}
\includegraphics[width=1.0\columnwidth]{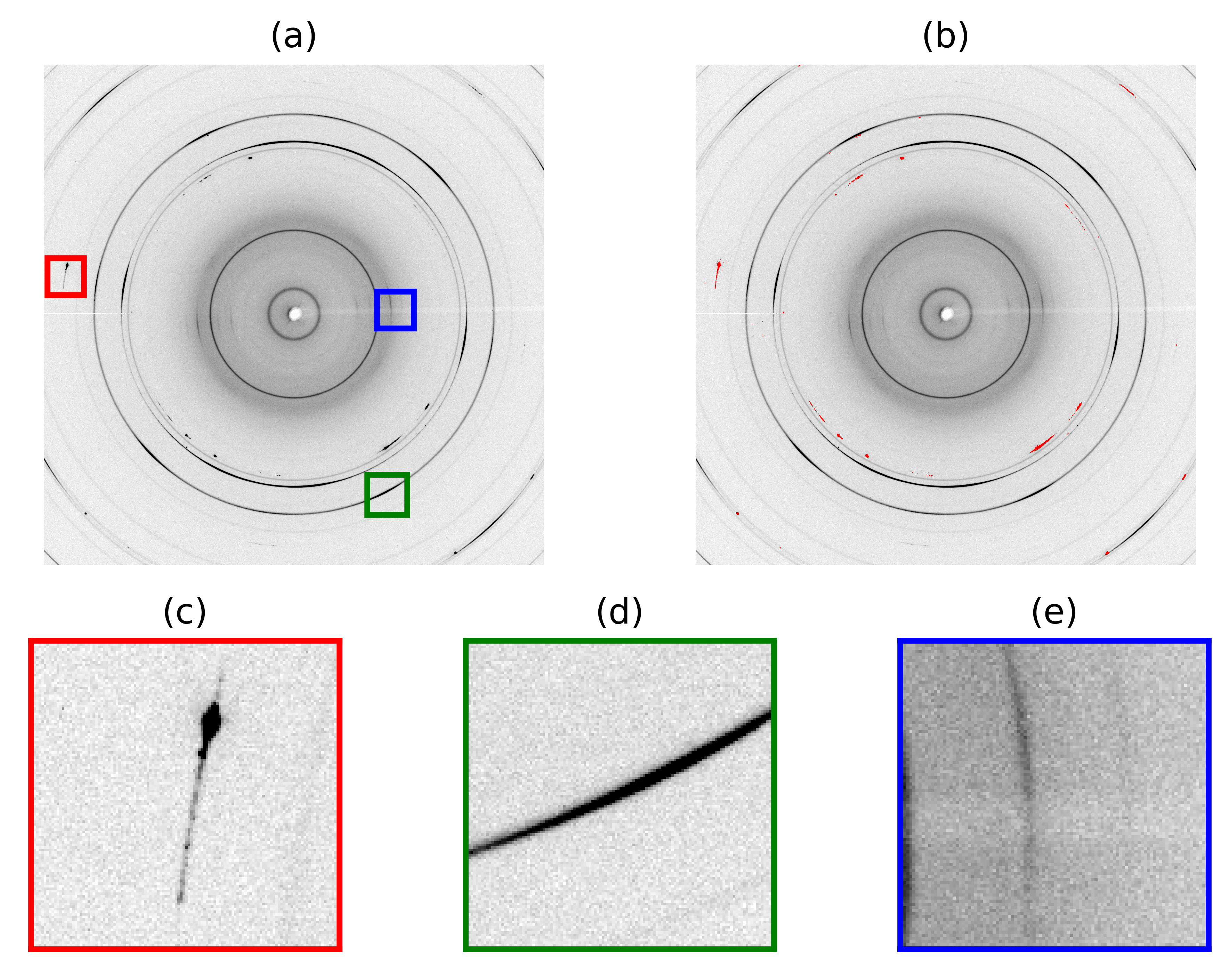}
\end{center}
   \caption{In (a), an experimental XRD image and (b) its masking result. The red, green and blue boxes in (a) show (c) single-crystal diffraction spots, (d) preferred orientation, and (e) two texture lines.}
   \label{xrd_image}
\end{figure}

In this study, various datasets of XRD images are utilized, which are obtained from different material compositions and under distinct experimental conditions (refer to Table \ref{datasets_table}), ensuring variability in XRD images. The experimental conditions can be subjected to, but are not limited to, different detector centers, wavelengths, and sample-to-detector distances. The Nickel83 dataset is collected during temperature ramping, and Battery-1, Battery-2, Battery-3, Battery-4, and Battery-5 datasets are collected during battery charging/discharging experiments, while the Perfect dataset contains images from different experiments with different materials. These images are high-resolution $2880\times2880$ pixel intensity arrays captured using a Varex XRD 4343CT area detector. Each XRD image may exhibit different characteristics, as illustrated in Figure \ref{xrd_image}.

The intensities of preferred orientation and single-crystal diffraction spots are often similar in magnitude, posing difficulty in their separation from an algorithmic perspective. Preferred orientation exhibits differential intensities around a powder diffraction ring, which are typically center-symmetric, which means an intense band at a particular azimuth angle accompanied by a similar intense band in the opposite direction or at 180$^{\circ}$. In contrast, single-crystal diffraction spots arise from large crystals, either from the materials in the sample holder or from single crystals formed during a reaction, e.g., a chemical synthesis process. Additionally, the overall pixel intensity scale of an XRD image is different from one experiment to another, depending on the inlet photon flux of the beam, sample scattering power, amount of sample in the beam path, total exposure time, and detector gain setting.

The Nickel83 dataset includes 11 images with powder diffraction rings and single-crystal diffraction spots. In contrast, the Battery-1 dataset has 11 XRD images with both single-crystal diffraction spots and powder rings displaying preferred orientation, making it more complex than the Nickel83 dataset. Due to the nature of the battery charging/discharging experiment, each image in the Battery-1 dataset exhibits distinct patterns that may represent different signals from the charge carrier. The Battery-2, Battery-3, Battery-4, and Battery-5 datasets are also associated with the charging/discharging experimental conditions and contain single-crystal spots and textures. The Perfect dataset, on the other hand, contains only the typical perfect powder rings. For creating the training and testing sets from these datasets, manual masking is carefully implemented for the single-crystal diffraction spots.

\subsection{U-Net Architecture}

\begin{figure*}
\begin{center}
\includegraphics[width=2.0\columnwidth]{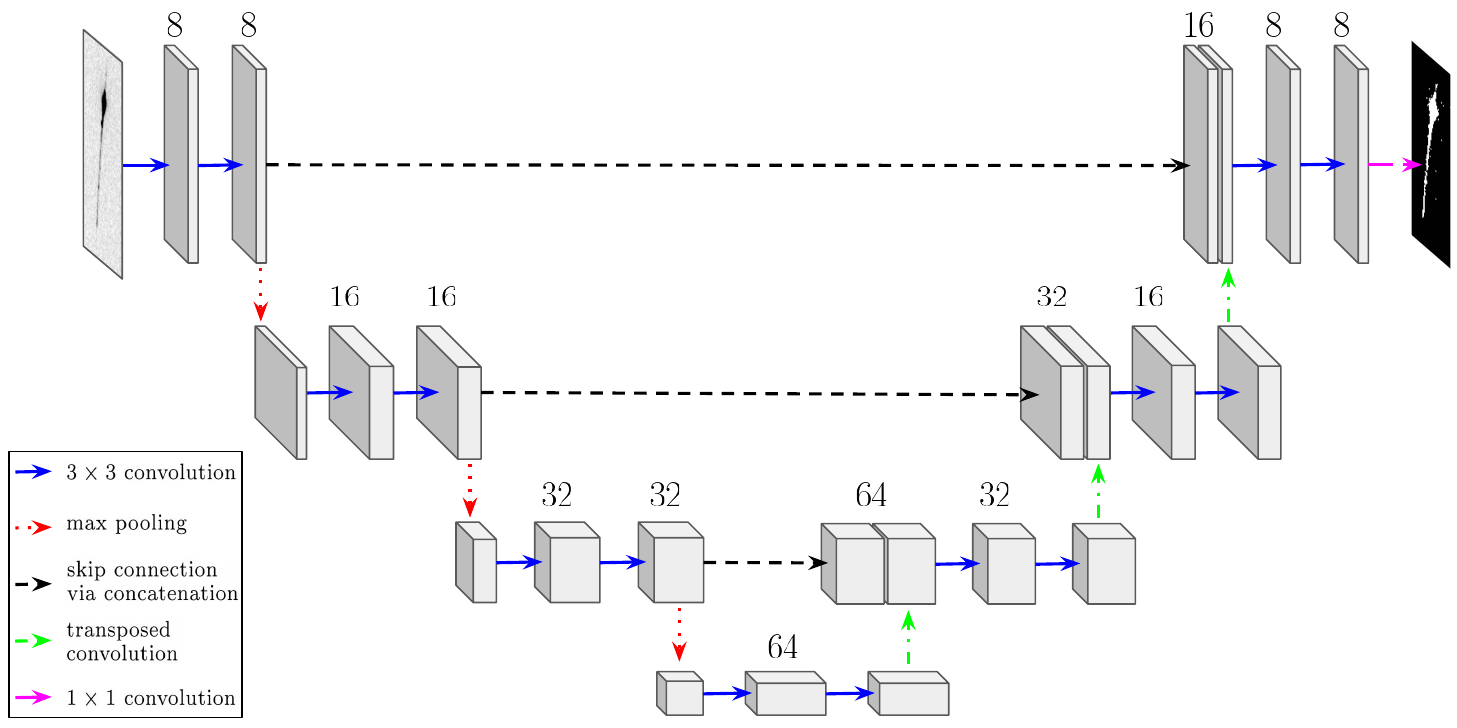}
\end{center}
   \caption{Schematic of a 4-layer U-Net with $8$ initial base channels and a convolutional channel growth/decay rate of $2$ between encoder/decoder layers. Pictured (left) is a typical grayscale input XRD image after cropping and (right) is the corresponding binary segmentation prediction mask output by the network.}
   \label{fig:unet}
\end{figure*}

U-Nets are a widely used and highly effective deep convolutional neural network with a symmetric encoder-decoder architecture originally developed for biomedical image segmentation \cite{ronneberger2015u}, consisting of a mirrored sequence of contractive and expansive operations at each predefined layer. The first-half encoder phase captures contextual information using back-to-back convolutional operations to extract image features, batch normalization and nonlinear activation (typically ReLU) to aid in gradient calculations and the optimization process \cite{ioffe2015batch}, and max-pooling operations to reduce spatial dimensions and capture higher level, more generalized features. The second-half decoder phase consists of familiar operations used in the encoder phase (dual convolutions, ReLU activation, and batch normalization) in addition to transposed convolutions to recover the original spatial dimensions and long-reaching skip connections \cite{drozdzal2016importance, long2015fully} in the form of channel-wise concatenations of intermediate feature maps between adjacent contractive and expansive phases. This decoder phase aims for the aggregation of multi-scale feature representation at different network stages, ultimately allowing for the learned features from the encoder to be projected into higher resolutions of the original image space for pixel-by-pixel semantic segmentation.

The performance of U-net on various applications depends heavily on the different hyperparameters. To aid in this hyperparameter search, this study employs the Deep Learning for Scientific Image Analysis (DLSIA) software suite \cite{roberts2023dlsia}, whose API provides complete freedom to create and deploy custom-sized and morphologically diverse U-Nets by specifying the following parameters:
\begin{enumerate}
    \item depth $d$: the number of layers in the U-Net, each consisting of dual convolutions and corresponding intralayer operations in both the encoder and decoder phases. Between each layer, there are $d-1$ mirrored max-pooling and up-convolutions,
    \item number of initial base channels $c_b$: the input data is mapped to this number of feature channels after the initial convolution, and
    \item growth rate $r$: the growth/decay rate of feature channels between encoder/ decoder layers. (Please note that DLSIA accommodates non-integer valued growth/decay rates.)
\end{enumerate}

\noindent Furthermore, DLSIA uses ReLU nonlinear activation and batch normalization as default after each convolution operation, but the user can apply any activation or normalization that complies with PyTorch syntax. Figure~\ref{fig:unet} shows a U-Net diagram representing one of the many tested in this study, with $d=4$ layers, $c_b=8$ base channels, and a growth rate of $r=2$. illustrating the sequence of operations and changes in channels and spatial dimensions in the contracting and expanding halves.

\begin{table*}[!t]
\begin{center}
\resizebox{2.0\columnwidth}{!}{\begin{tabular}{|l|l|ccccccc|}
\hline
\multirow{2}{*}{Sub-images} & \multirow{2}{*}{Dataset}  & \multicolumn{7}{c|}{Epochs}                                               \\
                            &                           & 10        & 20        & 30        & 50    & 75    & 100   & 200   \\
\hline \hline
\multirow{5}{*}{$128\times128$}    & Nickel83      & 88.8  $\pm$  6.8  & 91.9  $\pm$  4.7  & 90.7  $\pm$  5.3  & \textbf{91.6  $\pm$  3.0}  & 90.3  $\pm$  4.6  & 91.2  $\pm$  4.9  & 88.5  $\pm$  5.6 \\
                            & Battery-1     & 90.6  $\pm$  2.2  & 91.4  $\pm$  2.5  & 91.4  $\pm$  4.4  & \textbf{92.7  $\pm$  3.0}  & 91.2  $\pm$  2.2  & 91.0  $\pm$  3.3  & 88.9  $\pm$  2.8 \\
                            & Battery-2     & 90.2  $\pm$  3.2  & 92.0  $\pm$  3.8  & 92.9  $\pm$  2.3  & \textbf{92.9  $\pm$  1.7}  & 92.4  $\pm$  1.6  & 92.9  $\pm$  1.7  & 90.1  $\pm$  2.1 \\ 
                            & Battery-3     & 91.0  $\pm$  1.0  & 91.4  $\pm$  1.5  & 91.4  $\pm$  1.4  & \textbf{91.1  $\pm$  1.4}  & 90.3  $\pm$  2.8  & 90.9  $\pm$  1.2  & 89.3  $\pm$  0.9 \\ 
                            & Battery-4     & 93.3  $\pm$  4.9  & 95.9  $\pm$  4.2  & 96.4  $\pm$  0.8  & \textbf{96.7  $\pm$  1.4}  & 96.0  $\pm$  1.8  & 96.2  $\pm$  2.7  & 93.5  $\pm$  3.5 \\ 
\hline
\multirow{5}{*}{$256\times256$}    & Nickel83      & 87.5  $\pm$  4.5  & 89.7  $\pm$  1.6  & 90.9  $\pm$  3.1  & 88.5  $\pm$  4.5  & \textbf{92.7  $\pm$  2.0}  & 90.9  $\pm$  4.4  & 91.1  $\pm$  1.6 \\
                            & Battery-1     & 87.8  $\pm$  7.3  & 91.6  $\pm$  2.7  & 93.8  $\pm$  1.2  & 92.8  $\pm$  3.3  & \textbf{94.0  $\pm$  1.5}  & 93.4  $\pm$  2.8  & 92.5  $\pm$  1.6  \\ 
                            & Battery-2     & 84.2  $\pm$  11.8  & 89.7  $\pm$  4.3  & 93.4  $\pm$  2.1  & 92.3  $\pm$  2.1  & \textbf{93.2  $\pm$  1.2}  & 93.1  $\pm$  1.9  & 91.5  $\pm$  1.7 \\ 
                            & Battery-3     & 86.0  $\pm$  15.7  & 90.8  $\pm$  1.0  & 91.8  $\pm$  0.6  & 90.7  $\pm$  2.6  & \textbf{92.1  $\pm$  0.6}  & 92.0  $\pm$  1.0  & 91.1  $\pm$  1.3 \\ 
                            & Battery-4     & 87.8  $\pm$  7.6  & 95.4  $\pm$  2.8  & 96.4  $\pm$  2.0  & 93.4  $\pm$  9.5  & \textbf{97.0  $\pm$  1.4}  & 96.9  $\pm$  2.2  & 96.1  $\pm$  1.2 \\ 

\hline
\multirow{5}{*}{$512\times512$}    & Nickel83      & 68.4  $\pm$  18.0  & 81.8  $\pm$  11.1  & 87.1  $\pm$  4.5  & \textbf{89.7  $\pm$  4.6}  & 88.1  $\pm$  3.7  & 88.7  $\pm$  2.0  & 86.9  $\pm$  4.9 \\
                            & Battery-1     & 83.3  $\pm$  12.8  & 87.3  $\pm$  9.3  & 90.2  $\pm$  2.2  & 92.1  $\pm$  3.0  & 93.2  $\pm$  4.1  & \textbf{94.8  $\pm$  1.4}  & 87.1  $\pm$  9.3 \\ 
                            & Battery-2     & 80.2  $\pm$  11.1  & 87.1  $\pm$  2.0  & 86.9  $\pm$  2.4  & 91.8  $\pm$  2.9  & 92.9  $\pm$  2.2  & \textbf{94.2  $\pm$  0.5}  & 90.0  $\pm$  6.3 \\ 
                            & Battery-3     & 75.5  $\pm$  16.9  & 89.6  $\pm$  2.0  & 90.0  $\pm$  1.0  & 90.9  $\pm$  1.3  & 91.2  $\pm$  1.6  & \textbf{91.7  $\pm$  0.6}  & 89.2  $\pm$  6.9 \\ 
                            & Battery-4     & 50.4  $\pm$  27.4  & 79.2  $\pm$  18.4  & 88.3  $\pm$  8.8  & 96.0  $\pm$  3.0  & 96.1  $\pm$  2.3  & \textbf{97.3  $\pm$  1.2}  & 96.0  $\pm$  0.5 \\ 

\hline
\multirow{5}{*}{$1024\times1024$}  & Nickel83      & 57.5  $\pm$  11.9  & 53.6  $\pm$  26.4  & 64.0  $\pm$  21.8  & 71.4  $\pm$  16.8  & 75.9  $\pm$  9.7  & \textbf{78.0  $\pm$  7.3}  & 77.5  $\pm$  15.5 \\ 
                            & Battery-1     & 88.8  $\pm$  7.1  & 82.7  $\pm$  13.3  & 76.8  $\pm$  16.1  & 74.8  $\pm$  18.3  & 82.7  $\pm$  4.7  & 86.2  $\pm$  7.0  & \textbf{91.6  $\pm$  1.1} \\ 
                            & Battery-2     & 91.9  $\pm$  5.0  & 88.2  $\pm$  12.0  & 75.2  $\pm$  11.3  & 77.0  $\pm$  21.0  & 81.3  $\pm$  8.9  & 84.9  $\pm$  9.2  & \textbf{90.6  $\pm$  3.6} \\ 
                            & Battery-3     & 87.6  $\pm$  7.0  & 85.6  $\pm$  5.2  & 81.3  $\pm$  6.2  & 76.9  $\pm$  30.4  & 85.8  $\pm$  7.1  & 89.1  $\pm$  2.2  & \textbf{89.8  $\pm$  0.9}  \\ 
                            & Battery-4     & 52.3  $\pm$  21.5  & 42.4  $\pm$  16.4  & 53.6  $\pm$  21.9  & 63.6  $\pm$  28.1  & 76.7  $\pm$  17.8  & 80.8  $\pm$  15.0  & \textbf{89.0  $\pm$  7.5} \\ 
\hline
\end{tabular}}
\end{center}
\caption{True positive rate results in percentage over several training epochs of the test dataset. Bold texts indicate the best prediction across epochs.}
\label{epoch_table}
\end{table*}

\subsection{Training and Evaluation Details}\label{training_evaluation_details}

All training will utilize three random images from each dataset, while the remaining images will be used as the test dataset. To accommodate the large size of the XRD images and allow them to fit into the U-Net network, they are cropped into four separate datasets, each with smaller windows of sizes $128\times128$, $256\times256$, $512\times512$, and $1024\times1024$. Each dataset is cropped to allow a certain amount of overlap between adjacent sub-images, thus increasing the amount of original training data. The cropping process is carried out using the qlty Python package \cite{qlty2021}, which is also utilized for reassembling the cropped sub-images passed through the network back to the original image size. The overlapped regions were ignored in the reassembly process, thus alleviating edge effects. (For regression problems, the overlapped regions may typically be averaged, though this is not directly applicable for classification problems with integer class predictions.) During training, the cropped sub-images will be sorted according to the mean number of labeled pixels to ensure that labeled sub-images are prioritized due to data imbalance. Subsequently, 80\% of the cropped images will be randomly selected as training and validation sets, where the random selection contributes to the uncertainty quantification in Section \ref{optimization_subsection}. The training will involve shuffling the data for every epoch and using a batch size of 50.

The cross-entropy loss function was chosen to score the network predictions against the ground truth and was optimized using the ADAM optimizer \cite{kingma2014adam} to update the U-Net weights during training. The learning rate was set to a static $10^{-2}$ and training was performed on a single Nvidia A100 GPU with 40 GB or 80 GB. Due to the large size of $1024\times1024$ sub-images and the inability of the U-Net model to fit within a 40GB GPU, the training was done using an Nvidia A100 GPU with 80 GB. The remaining sub-image training was accomplished using an Nvidia A100 GPU with 40GB.

For evaluation purposes, the accuracy performance is measured by the recall and specificity scores \cite{yanxon2023artifact}. The true positive rate, or recall score, assesses a machine learning classifier's capacity to identify all masked pixels, while the true negative rate, or specificity score, evaluates the classifier's ability to detect all unmasked pixels. The primary evaluation metric will be the true positive rate, with the true negative rate reported as necessary. In cases where there is a class imbalance in the labeled data, as shown in Figure \ref{xrd_image}b, the recall or true positive rate score emphasizes accuracy for the infrequent class.

\section{Experiments}

\begin{figure*}[!t]
\begin{center}
\includegraphics[width=2.0\columnwidth]{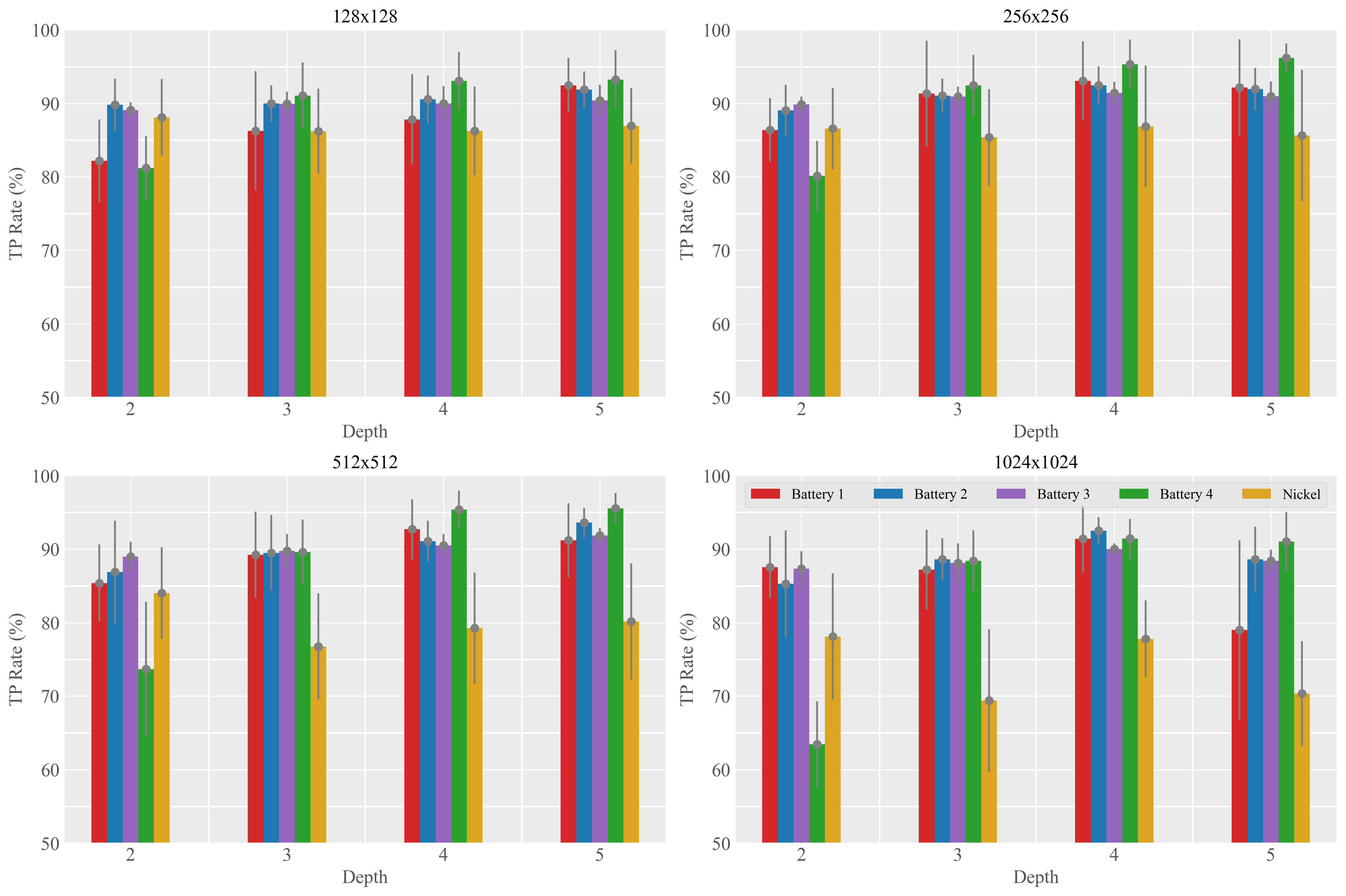}
\end{center}
   \caption{True positive rates applied to test datasets as a function of U-Net depth.}
\label{depth_figure}
\end{figure*}

In this section, we will discuss the procedure for tuning U-Net networks and examine the impact of incorporating additional data, such as the Perfect dataset and manipulated images, on performance. After training the model with the additional data, we will evaluate its performance on the Battery-5 dataset, which was not included in the training. Finally, we will examine the masking effects on the integrated 1D XRD pattern without masking and with the GSASII algorithm and U-Net masking procedures. 

\subsection{U-Net Optimization}\label{optimization_subsection}

The optimization process of the U-Net architecture will be demonstrated by examining the number of training epochs and the depth of the U-Net architecture, separately. In both cases, three images will be randomly chosen from each dataset to serve as the training dataset, while the remaining images will act as the test dataset. The training will be conducted 5 times to gather sufficient statistics to determine the uncertainty in the overall true positive rates for the test images. Training will be conducted using $c_b=8$ and a growth rate of 2. Hence, the number of learnable parameters for the U-Net varies based on the depth, with 6,562, 29,650, 121,394, and 487,154 parameters for depths 2, 3, 4, and 5, respectively.

The number of epochs necessary for training the U-Net architecture is examined using the Nickel83, Battery-1, Battery-2, Battery-3, and Battery-4 datasets. The images from these datasets are divided into smaller sub-images with resolutions of $128\times128$, $256\times256$, $512\times512$, and $1024\times1024$. The total numbers of training sub-images are 23,232, 5,292, 1,200, and 192. The model that is trained using these sub-images will be referred to as the 1X model for future reference. Table \ref{epoch_table} showcases the true positive rate accuracy of the U-Net networks against test datasets, as the training was performed for 5, 10, 20, 30, 50, 75, 100, and 200 epochs with U-Net depth of 4. It takes approximately 5 seconds per image for the prediction process. The results indicate that the predicted true positive rates stabilize at different epochs depending on the size of the sub-images. Additional training shows signs of overfitting.

\begin{table*}[t]
\begin{center}
\resizebox{1.3\columnwidth}{!}{\begin{tabular}{|l|c|c|c|c|c|c|}
\hline
 Dataset            & 0 & 90$^{\circ}$   & 180$^{\circ}$ & 270$^{\circ}$ & axis 0  & axis 1              \\
\hline \hline
 Nickel83             & 88.7(89.5)    & 82.1(88.8)  & 87.4(89.2)  & 82.3(88.9)  & 86.7(89.1)  & 86.8(89.0)  \\
 Battery-1          & 94.0(94.3)    & 70.9(93.9)  & 78.0(95.5)  & 73.5(95.5)  & 74.8(95.9)  & 74.1(94.9)  \\
 Battery-2          & 93.1(95.3)    & 86.0(94.6)  & 86.5(95.1)  & 83.0(95.0)  & 83.1(95.6)  & 81.3(95.0)  \\ 
 Battery-3          & 92.0(93.9)    & 85.2(93.7)  & 89.3(93.7)  & 85.6(93.9)  & 84.2(93.8)  & 86.3(93.8)  \\ 
 Battery-4          & 96.9(87.7)    & 53.4(94.5)  & 57.5(90.2)  & 54.5(94.1)  & 54.4(94.1)  & 68.8(92.1)  \\ 
\hline
\end{tabular}}
\end{center}
\caption{True positive rate prediction in percentage. The predictions are performed using the 1X(16X) U-Net model.}
\label{rotate_flip_table}
\end{table*}

The resolutions of $128\times128$, $256\times256$, $512\times512$, and $1024\times1024$ achieve their highest true positive rate predictions at epochs 50, 75, 100, and 200, respectively. Typically, the smaller resolutions reach their peak predictions earlier. The decline in accuracy is a result of overfitting occurring after the performance peak. The resolutions of $128\times128$, $256\times256$, and $512\times512$ demonstrate better prediction accuracy than the resolution of $1024\times1024$ across all datasets. This is likely due to the fact that the single-crystal diffraction spots are localized features in the XRD images. The resolution of $128\times128$ tends to perform better than the other resolutions at epoch $<$ 20, while resolutions of $256\times256$ and $512\times512$ can outperform $128\times128$ given more training epochs. Considering the lesser number of training sub-images and more training time required for the resolution of $512\times512$, the overall best performance for all datasets is demonstrated by the $256\times256$ resolution. As for the overall true negative rate, they are above 99.2\% and can reach as high as 99.9\% for all samples.

Given a $c_b$ value of 8 and a growth rate of 2, the U-Net model may gather only limited information from the resolution of $1024\times1024$ during convolution and max-pooling processes. The prediction resolution of $1024\times1024$ exhibits volatility at low epochs, as evidenced by the uncertainties. This behavior is attributed to the U-Net model being in the early stages of training, causing the model to inaccurately label most pixels as positive. Across all datasets, the overall true negative rate is approximately 20\% as it increases with more training epochs. The training can be further limited due to the size of the sub-images of 192. 


Additionally, we will examine the impact of U-Net depths on true positive rates using the Nickel83, Battery-1, Battery-2, Battery-3, and Battery-4 datasets. The U-Net architectures were trained for up to 200 epochs. Figure \ref{depth_figure} illustrates that the true positive rate improves with depth, but there is no substantial improvement beyond a depth of 4. The best results are achieved with resolutions of $128\times128$, $256\times256$, and $512\times512$ at a depth of 4, with $256\times256$ resolution having the best overall performance and reasonable uncertainty. The Nickel83 dataset generally performs worse than the others, as previously explained. These results confirm that the $1024\times1024$ resolution has reduced accuracy compared to other resolutions. Increasing U-Net complexity beyond these parameters did not improve performance significantly and can lead to overfitting. Therefore, the optimal training configuration is chosen to be around 75-100 epochs with $256\times256$ cropped images at a depth of 4 for the U-Net architecture. This setting balances accuracy and computational demand, and additional training will be conducted based on this configuration. 

\subsection{Additonal Training Images} 

In order to improve the U-Net prediction, we will incorporate additional XRD images such as rotated/flipped images and the Perfect dataset, which contains only perfect powder rings, simultaneously in the training. Initially, we will investigate rotated and flipped images in the Nickel83, Battery-1, Battery-2, Battery-3, and Battery-4 datasets using the optimally trained U-Net model in Subsection \ref{optimization_subsection}. The original images will be rotated by 90$^{\circ}$, 180$^{\circ}$, and 270$^{\circ}$, and flipped along axis 0 and axis 1. No other image manipulation, such as resizing, will be done, as it may produce unrealistic images, such as non-concentric powder rings. This process will increase the number of training sub-images of $128\times128$, $256\times256$, $512\times512$, and $1024\times1024$ resolutions from 23,232, 5,292, 1,200, and 192 to 381,004, 86,788, 19,680, and 3,148 sub-images, respectively. The model that is trained along with the rotated, flipped, and Perfect XRD images will be referred to as the 16X model for future reference. Table \ref{rotate_flip_table} presents the overall true positive rates obtained from the 16X U-Net model, which includes rotated and flipped XRD images. For comparison, the 1X U-Net model predictions are also displayed as a reference. The true positive rates for the manipulated XRD images predicted by the 1X model are lower than those for the non-manipulated XRD images, with a significant drop in the accuracy of Battery-4 predictions of about 50\%. However, the inclusion of the manipulated images leads to a significant improvement in prediction results, with some predictions even outperforming those for the non-manipulated XRD images. In the case of the Battery-4 dataset, there is a slight tradeoff, as the predicted results show about a 9.2\% lower in the overall true positive rates.

Furthermore, incorporating the Perfect dataset as a training dataset results in a significant reduction in false positives for the Nickel83, Battery-1, Battery-2, Battery-3, and Battery-4 datasets. Specifically, the average false positive counts per one XRD image change from approximately 46,000 to 16,000, 40,000 to 18,000, 26,000 to 30,000, 33,000 to 20,000, and 17,000 to 14,000, respectively. For the Perfect dataset, the average false positive counts per one XRD image decreased from 77,000 to 12,000 after the dataset is included in the training. Notably, both 1X and 16X U-Net models can accurately identify the absence of single-crystal spots in the Perfect dataset regardless of whether it is included in the training or not. The model's performance will be further evaluated on a previously unseen dataset, Battery-5. Strikingly, the 16X(1X) U-Net model achieves a true positive rate of 92.4\%(82.3\%) for the Battery-5 dataset. The average false positive of the 16X model is reduced by 34\% and 36\% in comparison to GSASII-algorithm masking and 1X model, respectively.

\subsection{XRD Data Analysis}

\begin{figure}[!t]
\begin{center}
\includegraphics[width=1.0\columnwidth]{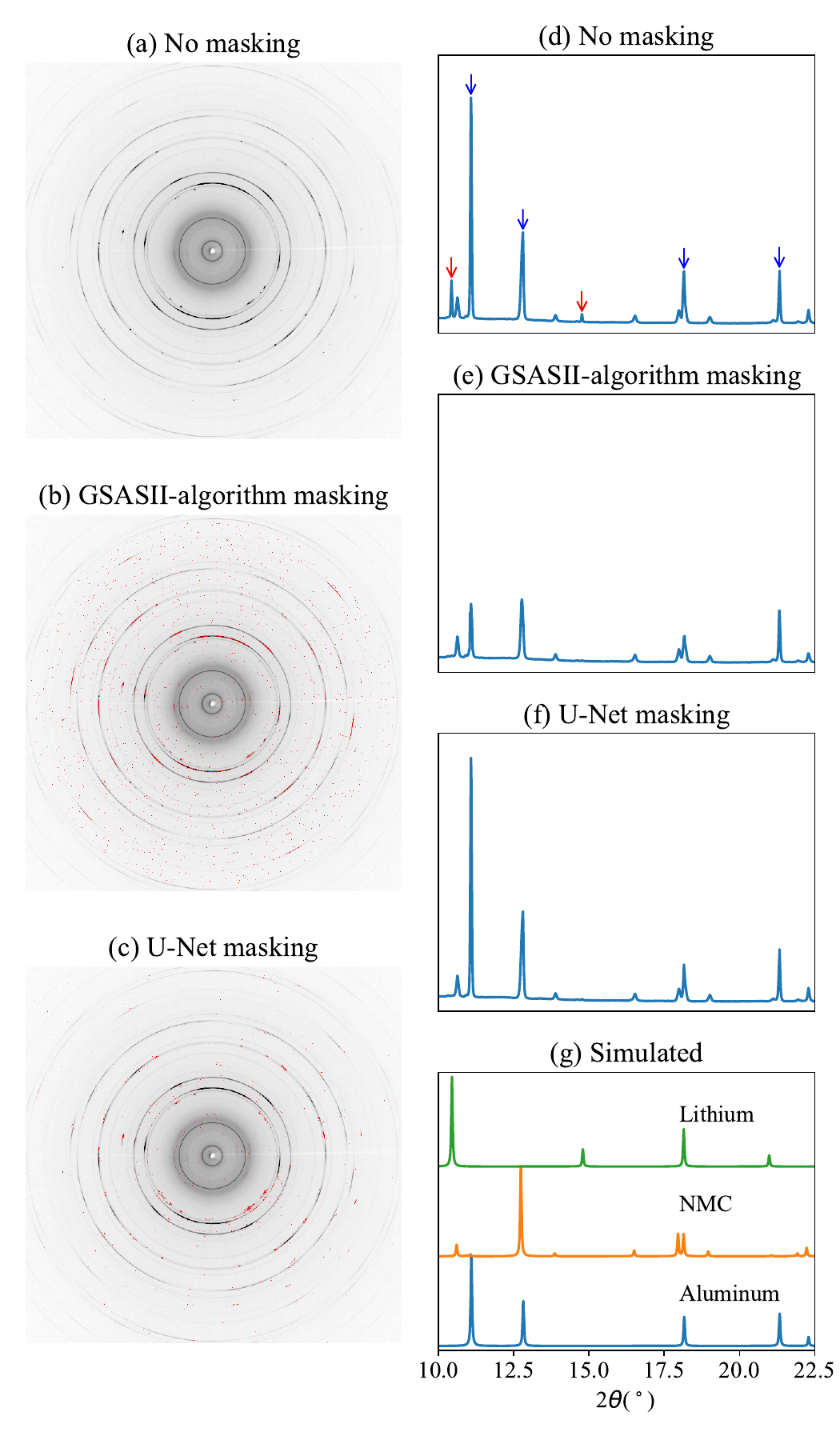}
\end{center}
   \caption{An XRD image of Battery-2 dataset (a) without mask, (b) with GSASII-algorithm mask, and (c) with U-Net mask. The corresponding integrated 1D patterns are displayed in (d), (e), and (f), respectively. (g) shows the calculated pattern from the three main phases in the sample, NMC, aluminum, and lithium. Some peaks from the aluminum and lithium phases are marked by blue and red arrows, respectively, in (d).}
\label{fig:1D_XRD}
\end{figure}

To demonstrate the importance of detecting and removing single-crystal diffraction spots or masking XRD images, the Battery-1 dataset will be utilized as an example. The dataset is obtained from an in situ synchrotron powder XRD experiment performed on a lithium-ion battery cell. As outlined in Table \ref{datasets_table}, the Battery-1 images exhibit various characteristics, single-crystal diffraction spots, powder rings with uniform intensities, and powder rings with non-uniform intensities indicating preferred orientation. The dataset captures diffraction signals from various phases, including cathode materials (nickel manganese cobalt oxide or NMC), lithium, aluminum, and graphite. Figure \ref{fig:1D_XRD} displays an XRD image of the Battery-1 dataset with (a) no mask, (b) GSASII-algorithm mask, and (c) U-Net mask along with their integrated 1D pattern (d) and (e), respectively. The diffraction spots are primarily attributed to the lithium phase (red arrows in Figure \ref{fig:1D_XRD}(d)), while the preferred orientation is a characteristic of the aluminum metal the integrated 1D XRD pattern (blue arrows in Figure \ref{fig:1D_XRD}(d)).

The integration process involves transforming an XRD image into a 1D XRD pattern that plots intensity against the diffraction angle, or 2$\theta$ values, which increase radially from the center of the XRD image. Prior to the integration process, each pixel of an XRD image is assigned a 2$\theta$ value by a calibration procedure. The integration process combines the counts of pixels with the same 2$\theta$ value and normalizes them to create a 1D pattern of intensity versus 2$\theta$ values. If a pixel is masked, its count is not included in the integration process, resulting in a 1D pattern that does not contain information about the masked pixel.

Both the GSASII algorithm and U-Net masking procedures eliminate the lithium diffraction signals in the 1D pattern, as indicated by the red arrows on the peaks. It is the intention to remove these peaks, which are caused by single-crystal diffraction spots, in order to ensure accurate 1D XRD analysis. However, the intense peak, that displays the preferred orientation (marked by the leftmost blue arrow), is greatly reduced due to the masking process of the GSASII algorithm to the XRD image, leading to an underestimation of the intensities for the aluminum phase. The U-Net-based masking, unlike the GSASII algorithm, does not lead to the underestimation of aluminum peak intensities in the integrated 1D pattern, as seen in the calculated/simulated aluminum pattern (Figure \ref{fig:1D_XRD}(g)). In consequence, the U-Net-based masked 1D pattern is more closely reflected in the calculated pattern, which assumes no preferred orientation. Furthermore, the GSASII algorithm masking process can be 2.2-18.4 times slower, making it unsuitable to work in conjunction with modern detectors that can capture data in sub-second intervals. In research using the Battery-2 dataset, the NMC cathode phase response to a charge-discharge process is the main focus. Thus, it is considered best practice to remove the spotty lithium phase and preserve the signatures of the aluminum phase. However, in situations where samples have multiple unknown phases, masking procedures can be used to group the peaks in the 1D XRD pattern and aid in correct phase identification.

\section{Conclusion}

Our study demonstrates the methods for tuning U-Net, incorporating manipulated (rotated and flipped) datasets, and utilizing U-Net masking before the integration process in the XRD analysis application. The results indicate that U-Net can achieve great predictions with over 92\% accuracy in the overall true positive rates when trained for 75-100 epochs at a resolution of $256\times256$, which is the optimal resolution for this application. This holds true for all datasets, despite the training dataset being small and imbalanced. The U-Net approach can provide precise predictions for datasets that were not included in the training process, with an overall true positive rate of 92.4\%, indicating a more robust model transferability. Furthermore, it was shown that the contrasting results among the integrated 1D XRD pattern without masking and those with the GSASII algorithm and U-Net masking. Given its ability to efficiently and accurately identify the relevant single-crystal diffraction spots, U-Net can maintain the overall integrity of the integrated 1D XRD pattern. Consequently, the U-Net masking approach proves to be a suitable method for identifying and separating artifacts in XRD images, paving the way for XRD experiment automation.

\noindent\textbf{Acknowledgements}
\noindent

This work is supported by the U.S. Department of Energy (DOE) Office of Science-Basic Energy Sciences award Collaborative Machine Learning Platform for Scientific Discovery. This research used resources of the Advanced Photon Source, a U.S. DOE Office of Science user facility at Argonne National Laboratory and is based on research supported by the U.S. DOE Office of Science-Basic Energy Sciences, under Contract No. DE-AC02-06CH11357.

The U.S. Government retains for itself, and others acting on its behalf, a paid-up nonexclusive, irrevocable worldwide license in said article to reproduce, prepare derivative works, distribute copies to the public, and perform publicly and display publicly, by or on behalf of the Government. The Department of Energy will provide public access to these results of federally sponsored research in accordance with the DOE Public Access Plan. http://energy.gov/downloads/doe-public-access-plan.

{\small
\bibliographystyle{ieee_fullname}
\bibliography{egbib}
}

\end{document}